# Zero-Shot Cross-Lingual Transfer Language Selection Using Linguistic Similarity

Juuso Eronen[a,*], Michal Ptaszynski[a] and Fumito Masui[a]

[a]Kitami Institute of Technology, 165, Koencho, Kitami, 090-0015, Hokkaido, Japan

## ARTICLE INFO

## ABSTRACT

We study the selection of transfer languages for different Natural Language Processing tasks, specifically sentiment analysis, named entity recognition and dependency parsing. In order to select an optimal transfer language, we propose to utilize different linguistic similarity metrics to measure the distance between languages and make the choice of transfer language based on this information instead of relying on intuition. We demonstrate that linguistic similarity correlates with cross-lingual transfer performance for all of the proposed tasks. We also show that there is a statistically significant difference in choosing the optimal language as the transfer source instead of English. This allows us to select a more suitable transfer language which can be used to better leverage knowledge from high-resource languages in order to improve the performance of language applications lacking data. For the study, we used datasets from eight different languages from three language families.

## 1. Introduction

As with any other supervised learning problem, the tasks in Natural Language Processing (NLP) require sufficiently large labeled datasets. What sets NLP apart from other fields is the presence of multiple languages the datasets can appear in. This means that in order to successfully train models for all of the world's 7,100 languages [1], one would need to annotate a dataset for each language. This is a very difficult and costly task [2, 3] and has led to a small number of high-resource languages to dominate the field [4, 5, 6]. This imbalance in the distribution of resources among languages calls for the need to develop technologies that would make model development for low-resource languages realistically feasible and efficient.

In order to address this problem, cross-lingual transfer been proposed as a solution. This means leveraging labeled data from high-resource languages in order to improve the performance on lower-resource languages [7, 8, 9, 10]. Particularly, the popularity of cross-lingual zero-shot learning, or training on one task/language and testing on a different task/language completely unknown to the model, has increased greatly in the recent years. Zero-shot learning has gained in popularity because it does not require any labeled data in the target language for training [11, 12]. Moreover, zero-shot cross-lingual transfer utilizes large pre-trained multilingual transformer models like Multilingual BERT [13] or XLM-RoBERTa [14]. These models are fine-tuned with training data in a language called the source language (usually a high-resource language) and then used to predict entries from other languages than that used in training, often with satisfying results [11, 15].

The choice of transfer language is usually done by intuition [16] or simply defaults to English, as is the case with popular multilingual benchmarks like XTREME [12] and XGLUE [17], even though there is no actual evidence backing up these choices. Furthermore, in a survey of 157 cross-lingual learning papers by Pikuliak et al. [18] they found out that English is used in 149 of those papers, followed by German with 82 papers in total. There has also been some attempts in developing a more systematic transfer language selection method [19]. However, this method requires training of a ranking model, limiting its use to the tasks and datasets used for training, making it unusable off-the-shelf for other applications.

Choosing the optimal language for cross-lingual transfer remains widely an understudied problem. Usually, the suitable source language candidate is decided experimentally or by pure intuition by the individual (researcher, or ML practitioner) based on their own theoretical knowledge and experience in the field. One option to select the source language would be by taking a look at languages that are from the same language group as the target language [20].

*Corresponding author

✉ eronen.juuso@gmail.com (J. Eronen)
ORCID(s): 0000-0001-9841-3652 (J. Eronen)

However, this does not guarantee that that the linguistic features shared between the two languages would be similar [21].

In order to contribute to the further understanding and solving this problem, we propose a method for choosing the source language for cross-lingual transfer. We show that there is a correlation between linguistic similarity and model performance, allowing us to select the best transfer language by comparing the source and target languages using different linguistic similarity measures. We also show that multilingual transformer models can be used to obtain good performance on the target language in a zero-shot learning setting. To select the optimal source language for transfer, we propose to quantify the features of languages to compute a metric that can be used in comparing the closeness of languages using their linguistic properties.

There are some existing metrics that use linguistic features in order to measure the linguistic distance between languages [22, 23, 24]. However, as these metrics simply take a handful or only a single linguistic feature into account, we propose a new linguistic similarity metric, which contains almost two hundred different features, based on the World Atlas of Language Structures (WALS) [25]. This allows us to not simply rely only on a single or a handful of features, but to have a more robust metric by better quantifying all aspects of the languages.

In this research we concentrated on three different Natural Language Processing tasks, namely, sentiment analysis, named entity recognition and dependency parsing. We used datasets from eight different languages, namely English, German, Danish, Polish, Croatian, Russian, Japanese and Korean. The languages were chosen as they have relatively high quality datasets available. Also, the languages represent different language families (English, German, Danish - Germanic; Polish, Russian, Croatian - Slavic; Japanese, Korean - Koreano-Japonic language family). This also gives us the opportunity to study the efficacy of cross-lingual transfer learning between and within language family groups.

In previous research [26] we showed that cross-lingual transfer performance correlates with the linguistic similarity of the prediction target language and the source language used for fine-tuning the models. Our hypothesis is that this is true for also other NLP tasks. In the experiments, we used multilingual transformer models, namely Multilingual BERT and XLM-RoBERTa, which were fine-tuned by using each of the languages as source and target. We calculated the linguistic similarity between all of our proposed languages using four different linguistic similarity metrics, EzGlot, eLinguistics, a quantified model based on WALS and averaged lang2vec. To demonstrate the effectiveness of our method, we then measured the correlation between the zero-shot cross-lingual transfer performance and linguistic similarity.

The paper outline is as follows. In Section 2 we go through all areas of previous research that are addressed in this paper. In Section 3 we describe all the tasks and datasets applied to this research and present their differences and features. In Section 4 we describe the applied multilingual transformer models and the linguistic similarity metrics used in this research. In Section 5 we describe our experiment workflow and go through all the results from the conducted experiments. In Section 6 we discuss the results in general and bring out the most interesting findings in relation to the research goals.

## 1.1. Contributions of This Study

The goal of this research is to develop a method for cross-lingual transfer language selection. Most often, the choice of a transfer source is made purely by relying on the practitioner's own judgement, using their accumulated experience on the field and theoretical knowledge or simply choosing a language from the same language family as the target [20]. The current methods have many problems as they are prone to bias from the practitioner and also completely unoptimized. In fact, one could even say that there is no systematic method usable off-the-shelf that could be used to determine, which languages should be considered as the cross-lingual transfer source.

We propose to investigate the possibility that different linguistic similarity metrics could be utilized when trying to find possible source language candidates for cross-lingual transfer also for other tasks than abusive language detection. We hypothesize that linguistic similarity correlates with cross-lingual transfer efficacy, meaning that by using more similar languages, we would be able to achieve higher model performance.

This research is was conducted in order to confirm the findings of our previous research [26] also with other Natural Language Processing tasks. We improved the calculation process of the linguistic similarity metric quantified from the World Atlas of Language Structures. This was done by selecting all of the features that would have a defined value for both languages in all possible language pairs instead of having to be shared between all of the languages, increasing the robustness of the metric. Also, we applied a linguistic similarity metric based on lang2vec by Littell et al. [27].

The main contributions of this work are as follows:

- We confirm the transfer language selection method based on linguistic similarity with multiple NLP tasks.

- We demonstrate the efficacy of two multidomain linguistic similarity metrics: improved quantified WALS and averaged lang2vec.

- We show that there is a significant difference in choosing an optimal transfer source language over English.

In practice, we propose to fine-tune cross-lingual pretrained transformer models, specifically mBERT and XLM-R, on three different Natural Language Processing tasks (sentiment analysis, named entity recognition and dependency parsing) using each of our proposed languages (English, German, Danish, Polish, Russian, Japanese, Korean) and then perform zero-shot prediction on the rest of the languages of the proposed set. We calculated the linguistic similarity between all of our proposed languages using four different linguistic similarity metrics, EzGlot, eLinguistics, quantified World Atlas of Language Structures and an averaged lang2vec proximity vectors. We then calculated the correlation between the zero-shot cross-lingual transfer performance and linguistic similarity to show the effectiveness of our method. A block-diagram of the system is shown in Figure 1.

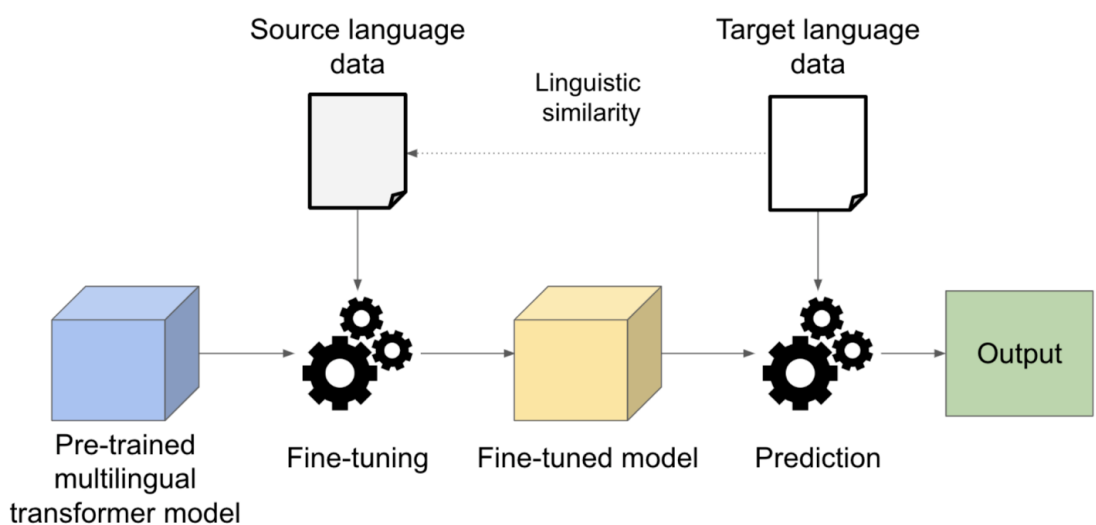

**Figure 1:** Block diagram of the proposed system

## 2. Previous Research

### 2.1. Measuring Linguistic Similarity

Already in 2006, the relation between the difficulty of language learning and the similarity of languages in general was discussed in a book by Ringbom [28]. The Finnish language scene was presented as an example in order to demonstrate the importance of cross-linguistic similarity in foreign language learning [29]. In short, he showed that Finnish-speaking Finns have a harder time learning English than Swedish-speaking Finns. The reason behind this being the closer relation between Swedish and English languages, giving an advantage to Swedish speakers when it comes to transferring the existing linguistic knowledge.

Cottorell et al. [30] showed that not every language is equally difficult to model. It was also shown by them that there is a correlation between the morphological richness of a language and the performance of the model. This means that the more complex the language is, the more difficult it becomes to model. This is hinting that more simple languages might not work so well when used as cross-lingual transfer sources for languages of higher complexity. This also implies that the direct relatedness (for example, language family) of languages should not be the only criteria in deciding the cross-lingual transfer source language as other features of the languages should also be thoroughly considered in order to find the most optimal transfer language.

There has been some research in attempting to quantify a linguistic similarity metric from different linguistic features. However, these metrics mostly commonly rely only on one or just a few different linguistic features. For example, by comparing the consonants contained in a predefined set of words while taking into account the order in which these consonants appear in the words, one can calculate a genetic proximity score between two languages. This is implemented as the eLinguistics [23] similarity metric. The metric makes it possible to get information about the direct relatedness of the compared languages. However, once the used languages start to become more and more distant, accidental similarities in consonants are introduced and there is a significant increase in the error rate. This is also acknowledged by the authors. Even though the metric is easy to calculate, it completely ignores all other kinds of linguistic features, for example, semantic, syntactic, or morphological.

Another method to calculate a similarity metric is to take a look at the vocabularies of two languages and concentrate on their similarity. EzGlot [24] uses lexical similarity as its basis for computation. The metric uses lexical similarity between the two compared languages while at the same time taking into account the amount of words the two languages are sharing with other languages. This allows for the calculation of similarity between the two languages in relation to the similarity with every other language.

Aggarwal et al. [22] proposed a linguistic similarity metric that utilizes multiple aspects of languages. Their metric, called STL, is based on Semantic, Terminological (lexical) and Linguistic (syntactic) similarity of languages. The method outperformed previous similarity metrics that concentrated only on one of the previously mentioned aspects [31, 32]. They noticed that the terminological measures showed a much higher contribution when compared to the other two features. However, in order to use the metric, the structure of the used vocabulary dataset needs to be in the form of a complex ontology. Due to this fact and because of the dataset only consisting of German, French, Italian, Dutch, Spanish and English, and due to the dataset used by the authors being no longer available, it was not feasible to use the metric as a part of this research.

The lang2vec developed by Littell et al. [27] is a database that represents languages as typological, phylogenetic, and geographical vectors, which are derived from a number of different linguistic resources, for example, WALS [25], PHOIBLE [33], Ethnologue [1], and Glottolog [34]. Each of these utilize multiple different features, making them more robust than the EzGlot or eLinguistics metrics. The lang2vec is a fully-fledged library that can be used to query for different linguistic features and to get pre-computed distances between languages, based on some typological information.

The World Atlas of Language Structures (WALS) project [25] consists of a database that catalogs phonological, word semantic and grammatical knowledge for 2,662 languages with almost two hundred different linguistic features from multiple domains. Using a linguistic similarity measure quantified from the WALS database into would allow a more robust method to measure similarity and would aid capturing all aspects of the languages instead of relying only on a single or a handful of linguistic features. Concentrating purely on using WALS to create a similarity metric would also preserve homogeneity and allow a more explainable and controllable implementation. In previous research [26] we proposed a novel linguistic similarity metric quantified from the WALS database. This metric proved to be more robust compared to the other metrics, at least for the applied abusive language detection task, as it was based on multiple kinds of linguistic features.

## 2.2. Transfer Language Selection

Selecting the optimal language for cross-lingual transfer remains mostly an unanswered question. Most of the time, the decision of which language to use as the transfer source comes up to the practitioner's consideration. This is usually done experimentally or by intuition [20, 35, 16] or by simply relying on English [36, 37, 38]. For example, in order to get a more successful transfer, Cotterell and Heigold [20] focused on using languages belonging to the same language family as the cross-lingual transfer target. However, even though the languages are part of the same language family, two languages could be very distant for example when looking at the complexity of grammar, which means that it does not guarantee them sharing the same linguistic features [21].

A common way for choosing the transfer language is to simply default to English. The reason being that it is the de-facto highest resource language available for most NLP tasks [4]. This is also the case with popular multilingual benchmarks like XTREME [12] and XGLUE [17]. Although, recently benchmarks like XTREME-R [39] have started to include cross-lingual training sets. Furthermore, in a survey of 157 cross-lingual learning papers by Pikuliak et al. [18] they found out that English was used in 149 papers, followed by German with 82 papers. Additionally, it has been shown that other languages than English, for example, German and Russian tend to work better as transfer sources [40].

Duong et al. [41] found out that choosing the transfer language based on language family is not optimal for many languages. For example, their experiments showed that the best source language for both Finnish and German is Czech, even though being from a different language family than the targets. They concluded that apparently, the best source language for cross-lingual transfer is not predictable from language family information. Instead, they proposed two methods for transfer language selection. The first being based on the Jensen-Shannon divergence between the distributions of parts-of-speech n-grams on a pair of languages. The second method was based on the word-order information feature in WALS. Both of these methods showed improvements over choosing English or a language from the same family as the target. They also experimented with using multiple source languages, which further improved the performance.

It has been shown [42, 43, 44] that transferring from many high-resource languages at the same time can yield higher results compared to selecting only a single language as the transfer source. However, these methods do not consider the actual relation between the source and the target languages and the amount of contribution of each of the languages to the total score. Also, Nooralahzadeh et al. [45] discovered that certain morphosyntactic features shared between languages tend to give a boost to cross-lingual transfer performance.

Lin et al. [19] developed a ranking method for possible transfer language candidates using the lang2vec metrics [27] together with dataset dependent features like word overlap and type-token ratio. they discovered that using both the dataset independent linguistic features and database dependent features to train the ranking model yields the best results. However, as their method requires training of the ranking model, it is dependent on the tasks and datasets used for training and is not usable out of the box for other applications.

In another study [38] it was shown that the transfer performance with English as the source correlates with the linguistic similarity metrics of lang2vec [27], meaning that target languages more similar to English yielded higher scores. They found out that similarity of syntactic structures especially play an important role in selecting the source language for tasks like parts-of-speech tagging (POS), named entity recognition (NER) and dependency parsing (DEP). They also discovered that the fine-tuning corpus size of the target language also makes a difference considering the cross-lingual transfer performance, especially for higher level tasks like question answering. However, their research concentrated only on using English as the source language and the capabilities of other languages as the transfer source were left completely unexplored.

Martinez et al. [46] found out that differences in language morphology in cross-lingual transfer generally lead to a higher loss than when transferring between languages with the same morphological typology. Furthermore, they showed that parts-of-speech tagging tends to be more sensitive towards changes in morphological typology compared to sentiment analysis, which seems to be more sensitive to variables related to the fine-tuning data and the transfer performance being generally harder to predict.

In their research, Gaikwad et al. [10] discovered that there could be a relation between cross-lingual transfer performance and language similarity. They classified entries in the Marathi language using multiple languages, specifically Bengali, Greek, English, Turkish and Hindi as cross-lingual transfer sources. Their results showed that the closest language of these to Marathi, Hindi, also had the highest performance. This hints that a solution to the problem of cross-lingual transfer language selection could be found with the aid of linguistic similarity.

In our previous research [26], we showed that there is a correlation between language similarity metrics and cross-lingual transfer efficiency, at least for offensive language identification. This allows for choosing of an ideal transfer language by using different metrics to compare the similarity languages without having to rely on one's intuition. We also showed that choosing a transfer language, for example, only by looking at the language family is not always the best option.

## 3. Tasks

In this research, we concentrate on three different NLP tasks. Sentiment analysis as a document classification task. Named entity recognition as a token classification task. And lastly, dependency parsing for understanding the importance of syntax and grammar in cross-lingual transfer.

We hypothesize that the zero-shot cross-lingual transfer performance correlates with the linguistic similarity of the source and target languages. In order to confirm our hypothesis, we used datasets from eight different languages, namely English, German, Danish, Polish, Russian, Croatian, Japanese and Korean for all of the tasks. We chose these languages as they had high quality datasets compared to other options and because the languages represent three different language families (English, German, Danish - Germanic; Polish, Russian, Croatian - Slavic; Japanese, Korean - Koreano-Japonic language family). This also gives us the opportunity to study the efficacy of cross-lingual transfer learning between and within different language family groups.

## 3.1. Sentiment Analysis

In the field of NLP, sentiment analysis is one of the most active research areas [47]. The recent research in sentiment analysis, as with many other NLP tasks, has mainly focused on using deep neural networks and pretrained language models [48, 49, 50, 51, 52]. The popularization of multilingual transformer models has made it possible to utilize cross-lingual transfer in order to train models for low-resource languages.

Rasooli et al. [53] used a set of 16 languages from different language families, namely Indo-European, Turkic, Afro-Asiatic, Uralic, and Sino-Tibetan, to learn a sentiment analysis model. Their experiments showed that for most target languages the best result can be obtained by leveraging from multiple source languages at the same time. Also, datasets of a similar genre and domain tended to yield higher results when compared to out-of-domain and dissimilar genres.

Pelicon et al. [54] used zero-shot cross-lingual transfer to classify Croatian news articles with an mBERT model fine-tuned using Slovene data with good results. In addition, Kumar et al. [55] used XLM-R and performed cross-lingual transfer from English to Hindi. Their model compared favorably to the used benchmarks and gives an effective solution to the analysis of sentiments in a resource-poor scenario.

The majority of the sentiment analysis datasets used in this research consists of product reviews, as we attempted to keep the domain the same throughout the languages. However, for some languages, we were unable to find such data, most notably Croatian, which consists of news articles. We also had to adjust the labels of some of the datasets so that they would match among all of the languages. Training and evaluation splits were retained from original datasets if possible, otherwise datasets were split to 80% training and 20% evaluation.

For this research we used the Multilingual Amazon Reviews Corpus [56], which covers English, Japanese and German. The dataset contains over 200,000 reviews for each language collected between 2015 and 2019. The reviews are labeled from one to five stars. However, as the other datasets used in this research used a two-point scale (positive, negative), we adjusted the labels accordingly (positive: 5 and 4 stars, negative: 2 and 1 stars).

For Danish, we used a dataset containing almost 45,000 reviews crawled from Trustpilot by Alessandro Gianfelici[1]. For Polish, we used the PolEmo 2.0 corpus [57]. This dataset contains over 8,000 reviews from the domains of medicine, hotels, products and school. For both of these datasets, we also had to adjust the labels of this dataset to a two-point scale similarly to the Amazon Reviews dataset.

The Russian dataset used in this research was a product review dataset by Smetanin et al. [58]. The dataset consists of 90,000 automatically labeled reviews on the topic "Women's Clothes and Accessories", split evenly among three classes (positive, neutral, negative). The Croatian dataset is the same used by Pelicon et al. [54], containing around 2,000 news articles. The articles were collected from 24sata, one of the leading Croatian media companies. The annotations were done by 6 people using a five-level Likert scale. The annotations were later adjusted to a three-point scale by the authors. For the purpose of our experiments, in case of both datasets, we left out the neutral reviews in order to binarize the labels.

The Korean dataset used in this research was Naver sentiment movie corpus v1.0[2]. The dataset consists of Naver Movie reviews, with 100,000 positive and negative samples. The reviews were originally rated from one to ten, but the creators binarized the dataset prior to publishing.

## 3.2. Named Entity Recognition

The research on Named Entity Recognition (NER) has also shifted towards using Deep Neural Networks and most recently, pretrained transformer models [59, 60, 61]. Cross-lingual transfer has also been applied to NER in multiple research. Fritzler et al. [62] used a metric-learning method to at the time outperform a state-of-the-art recurrent neural network method and showed to be capable in both few-shot and zero-shot settings. Moon. et al. [63] used multilingual BERT to fine-tune a NER model in multiple languages and showed it to be more effective than a model fine-tuned only on a single language. This demonstrates that the model can leverage knowledge from other languages in order to improve its performance on one.

Hvingelby et al. [64] presented a Danish NLP resource based on the Danish Universal Dependencies treebank and showed that transferring from other Germanic languages, especially from English and Norwegian, to Danish can yield good results when using mBERT. However, using other Germanic languages in addition to Danish did not give any better results compared to fine-tuning only with Danish in their case.

---

[1]https://github.com/AlessandroGianfelici/danish_reviews_dataset
[2]https://github.com/e9t/nsmc

Entity projection [65, 66] has been used to generate pseudo-labeled datasets for low-resource NER datasets with the help of parallel corpora. However, it has been shown by Weber and Steedman [67] that entity projection can be outperformed by cross-lingual transfer and XLM-RoBERTa. The reason behind this could be explained by the discovery by Lauscher et al. [38], who showed that transfer performance with English as the source correlates with the similarity of the languages when dealing with a NER task.

In this study, we used the WikiANN [68] multilingual NER dataset also used by XTREME benchmark [12] for all of the proposed languages. WikiANN consists of Wikipedia articles annotated with LOC (location), PER (person), and ORG (organisation) NER tags. We used the version by Rahimi et al. [69], which has a balanced train, development, and test splits and supports 176 of the 282 languages from the original WikiANN corpus.

### 3.3. Dependency Parsing

Cross-lingual transfer in dependency parsing (DEP) has been studied for some time before the advent of multilingual transformer models [70, 71, 72, 73]. These studies mainly used deep neural network-based methods on parallel corpora. The research by Duong et al. [41] discussed earlier in Section 2.2 was also conducted on a dependency parsing task. Instead of using parallel corpora, their research was built around syntactic cross-lingual word embeddings [74] trained over POS contexts to emphasize similarity.

Multilingual transformer models have also seen success in the dependency parsing task [15, 75, 76]. Most notably, in their study, Lauscher et al. [38] discovered that structural and syntactic similarities between languages are the most determining factor when it comes to the success of cross-lingual transfer for lower-level tasks like POS-tagging and DEP.

The dataset used for all of the proposed languages in this study was the Universal Dependencies v2 [77], a widely used resource in NLP as well as in linguistic research. The dataset was also used in the XTREME [12] benchmark and in the research by Lauscher et al. [38] described earlier. Universal Dependencies is a framework for a consistent annotation of grammar, including parts-of-speech, morphological features, and syntactic dependencies across a total of more than 100 languages.

## 4. Methods

### 4.1. Models

For the experiments we used two pre-trained multilingual transformer models. The experiments were carried out in a zero-shot cross-lingual setting [78], meaning that the fine-tuning is done using only data from another language than the target language.

**Multilingual BERT** (mBERT) [13] is the multilingual version of BERT, which stands for Bidirectional Encoder Representations from Transformers. It is based on an attention mechanism called the Transformer [79] that learns contextual relations between words (or sub-words) in text. One of the features transformer models introduced is the capability to read text input in both directions at once, instead of being able to only read it sequentially from left-to-right or right-to-left. Taking advantage of this bidirectional capability, BERT is pre-trained on two NLP tasks, Masked Language Modeling and Next Sentence Prediction. The objective of Masked Language Modeling is to mask a word in a sentence and have the algorithm predict based on the word's context what word has been hidden. In Next Sentence Prediction, the algorithm takes two masked sentences and needs to predict if they have a sequential connection or not.

Although mBERT has not been trained using any cross-lingual data, it has showed cross-lingual capabilities and had good results in many cross-lingual tasks [80]. This also includes various zero-shot transfer tasks. Multilingual BERT has even been shown to outperform the usage of various cross-lingual embeddings [15]. This ability to generalize could come from having word pieces used in all languages, for example, numbers, URLs, etc, mapped to a shared space. This in turn forces the co-occurring pieces to also be mapped to a shared space, thus spreading the effect to other word pieces, until different languages are close in a shared space [11].

**XLM-RoBERTa** (XLM-R) [14] is a multi-lingual transformer model, also trained with the Masked Language Model objective. XLM-R is trained on around a total of 2.5tb of CommonCrawl data in one hundred different languages. The model is trained in the same way as the monolingual RoBERTa [81]. This means, that the only objective in its pre-training is Masked Language Modeling. The model is not trained on the Next Sentence Prediction task like BERT or using the parallel Translation Language Model objective of XLM.

XLM-R has been shown to outperform both mBERT and XLM on a many cross-lingual benchmarks, including zero-shot cross-lingual transfer tasks [82]. It has also been shown to perform well on low-resource languages. A

**Table 1**
eLinguistics metric between all applied languages

|          | Danish | English | German | Croatian | Polish | Russian | Japanese | Korean |
|----------|--------|---------|--------|----------|--------|---------|----------|--------|
| **Danish**   | 0.00   | 20.60   | 38.20  | 66.20    | 68.20  | 66.20   | 95.20    | 97.20  |
| **English**  | 20.60  | 0.00    | 30.80  | 60.30    | 66.90  | 60.30   | 88.30    | 90.00  |
| **German**   | 38.20  | 30.80   | 0.00   | 64.50    | 68.10  | 64.50   | 87.40    | 95.50  |
| **Croatian** | 66.20  | 60.30   | 64.50  | 0.00     | 10.70  | 5.60    | 90.70    | 87.20  |
| **Polish**   | 68.20  | 66.90   | 68.10  | 10.70    | 0.00   | 5.10    | 93.30    | 89.50  |
| **Russian**  | 66.20  | 60.30   | 64.50  | 5.60     | 5.10   | 0.00    | 93.30    | 89.50  |
| **Japanese** | 95.20  | 88.30   | 87.40  | 90.70    | 93.30  | 93.30   | 0.00     | 88.00  |
| **Korean**   | 97.20  | 90.00   | 95.50  | 87.20    | 89.50  | 89.50   | 88.00    | 0.00   |

notable feature of XLM-R is that it can also match the performance of to state-of-the-art monolingual models, which demonstrates that it is possible to create multilingual models without sacrificing per-language performance in a monolingual setting [14], most likely thanks to the sheer amount of data used in the pre-training.

## 4.2. Linguistic Similarity Metrics

To be able to calculate the correlation between cross-lingual zero-shot transfer performance and language similarity for the proposed tasks, we needed a way to quantify the aspects of all of the languages in our proposed set, specifically, a language similarity metric. We utilized four language similarity measures, eLinguistics [23], EzGlot [24], the multidomain metric we quantified from the linguistic features presented in WALS [25] and averaged genetic, geographic, syntactic, inventory, phonological and featural metrics from lang2vec [27]. We propose that linguistic similarity metrics could be utilized when trying to find optimal source language candidates for cross-lingual transfer. We hypothesize that linguistic similarity correlates with cross-lingual transfer efficacy, meaning that by using more similar languages, we would be able to achieve higher model performance.

**eLinguistics** [23] works by calculating a genetic proximity value for a pair of languages based on the use of phonetic consonants. The score is calculated by taking a predefined word set and comparing the consonants contained in these words. The method also takes into account the order of the consonants. This way, it is possible to get information regarding the closeness of the phonetics of the pair of languages set for comparison. The assessment of the relationship of the consonants is based on the research done by Brown et al [83].

Even though completely disregarding semantic, morphological, and syntactic similarity and being very simple in formulation, the similarity values produced by the method seemed to be in line with our expectations and the two multidomain metrics (WALS, lang2vec) used in this research. However, as the distance between the two compared languages increased, the method seemed to become increasingly more prone to errors. This is due to the surging amount of accidental similarities in consonants. The similarity measure can be accessed from a web service[3]. The similarity values between our proposed languages are shown in Table 1.

**EzGlot** [24] is based on the similarity of vocabularies, or lexical similarity, of the two compared languages. EzGlot's similarity metric is computed by taking the lexical similarity between the two compared languages, while in addition taking into account the number of words the pair of languages also have in common every other language. This makes it possible to compute a similarity measure for a pair of languages in relation to their closeness with every other language. Also, due to including the calculation of the number of words the languages share with all other languages, the similarity measure becomes asymmetric between every pair of languages. This also supports studies stating that mutual language intelligibility is being considered asymmetric as well [84, 21].

A pre-computed language similarity matrix and the formula for its computation can be found on the EzGlot similarity metric project's web page[4]. However, the usability of the metric is hindered by the high amount of missing values in the similarity matrix. For example taking a look at Japanese, which is one of the languages utilized in our experiments, over half of the values are missing for our proposed languages. Also, the authors of the similarity measure do not give away their data source. This means that we are unable to say anything regarding the quality of the computations. This also makes it more difficult to fill in the missing values to the similarity matrix. We extracted the similarity values from the EzGlot's similarity matrix for the proposed languages. These values are presented in Table 2.

---

[3]http://www.elinguistics.net/Compare_Languages.aspx
[4]https://www.ezglot.com/most-similar-languages.php

**Table 2**
EzGlot metric between all of the proposed languages

|  | Danish | English | German | Croatian | Polish | Russian | Japanese | Korean |
|---|---|---|---|---|---|---|---|---|
| Danish | 100 | 9 | 17 | N/A | 13 | N/A | N/A | 9 |
| English | 6 | 100 | 28 | 6 | 19 | 14 | 7 | 26 |
| German | 6 | 15 | 100 | 4 | 8 | 4 | N/A | 5 |
| Croatian | N/A | 4 | 5 | 100 | 14 | 9 | N/A | 5 |
| Polish | 6 | 12 | 9 | 14 | 100 | 15 | N/A | 5 |
| Russian | N/A | 11 | 7 | 11 | 19 | 100 | N/A | 11 |
| Japanese | N/A | 2 | N/A | N/A | N/A | N/A | 100 | 8 |
| Korean | 1 | 5 | 2 | 1 | 1 | 3 | 4 | 100 |

**Table 3**
Averaged lang2vec metric between all of the proposed languages

|  | Danish | English | German | Croatian | Polish | Russian | Japanese | Korean |
|---|---|---|---|---|---|---|---|---|
| Danish | 0.000 | 0.511 | 0.487 | 0.550 | 0.565 | 0.597 | 0.694 | 0.691 |
| English | 0.511 | 0.000 | 0.352 | 0.578 | 0.486 | 0.488 | 0.635 | 0.578 |
| German | 0.487 | 0.352 | 0.000 | 0.550 | 0.470 | 0.471 | 0.594 | 0.579 |
| Croatian | 0.550 | 0.578 | 0.550 | 0.000 | 0.513 | 0.505 | 0.709 | 0.699 |
| Polish | 0.565 | 0.486 | 0.470 | 0.513 | 0.000 | 0.344 | 0.624 | 0.619 |
| Russian | 0.597 | 0.488 | 0.471 | 0.505 | 0.344 | 0.000 | 0.589 | 0.585 |
| Japanese | 0.694 | 0.635 | 0.594 | 0.709 | 0.624 | 0.589 | 0.000 | 0.518 |
| Korean | 0.691 | 0.578 | 0.579 | 0.699 | 0.619 | 0.585 | 0.518 | 0.000 |

**Averaged lang2vec** is calculated from genetic, geographic, syntactic, inventory, phonological and featural metrics of lang2vec. Lang2vec [27] is a database that provides vector identifications of languages based on different linguistic features based on various linguistic resources like WALS [25], PHOIBLE [33], Ethnologue [1], and Glottolog [34]. The lang2vec is a fully-fledged library that can be used to query for different linguistic features and to get pre-computed genetic, geographic, syntactic, inventory, phonological and featural distances between languages. In order to use lang2vec as a multidomain linguistic similarity metric, we used an average value of these six categories.

The method is based on multiple types of linguistic features, making it naturally more robust than EzGlot or eLinguistics similarity metrics, which only rely on a single kind of linguistic feature each. The method also uses a larger amount of data compared to the previously described metric based on WALS. Additionally, lang2vec deals with the missing values in linguistic resources by predicting them [85]. However, due to being based on multiple sources, the heterogeneous nature of the method brings up many questions. For example, there might be incoherence as we do not know how features are selected from different sources and how they are weighted. Also, the features are one-hot encoded which causes a complete loss of ordinality between feature values. Additionally, using geographical information as one of the vectors seems questionable as it was shown to be unreliable when predicting similarity of languages [86]. The averaged distance matrix for lang2vec is shown in Table 3.

**Quantified World Atlas of Language Structures** is a similarity metric developed by us in previous research [26]. It is based on The World Atlas of Language Structures (WALS) [25], which is a massive language database that records phonological, word semantic and grammatical information for a total of 2,662 languages from over 200 different language families. There are 192 different linguistic features in the database currently (May 2022). However, many of the linguistic features are missing for of the available languages. For example, one of the most extensively documented language, English, has about 150 features documented in the database. This amount rapidly decreases for languages studied less. Taking Danish as an example, it only 58 features documented[5]. Considering every language and all of the features, this adds up to over 58,000 data points in total in the WALS database. This means the whole database is only approximately 12% populated, meaning a vast majority of the information is missing. Also many major and widely studied languages are missing many features. For example, 25% of all of the features are missing for English. These missing values and the sparsity of the data is the main point of concern when quantifying the WALS database into a linguistic similarity metric as using lesser known and not so widely studied languages means having less common features among them.

---

[5]Some even less studied languages have an even smaller number of features documented, e.g. Chuj language, spoken in Guatemala, has only 29, while the Indonesian Kutai language has only a single feature documented.

**Table 4**
Quantified WALS metric between all of the proposed languages

|          | Danish | English | German | Croatian | Polish | Russian | Japanese | Korean |
|----------|--------|---------|--------|----------|--------|---------|----------|--------|
| Danish   | 0.000  | 0.109   | 0.140  | 0.167    | 0.197  | 0.155   | 0.236    | 0.202  |
| English  | 0.109  | 0.000   | 0.136  | 0.179    | 0.164  | 0.141   | 0.252    | 0.209  |
| German   | 0.140  | 0.136   | 0.000  | 0.221    | 0.196  | 0.140   | 0.248    | 0.225  |
| Croatian | 0.167  | 0.179   | 0.221  | 0.000    | 0.160  | 0.080   | 0.272    | 0.229  |
| Polish   | 0.197  | 0.164   | 0.196  | 0.160    | 0.000  | 0.097   | 0.249    | 0.210  |
| Russian  | 0.155  | 0.141   | 0.140  | 0.080    | 0.097  | 0.000   | 0.225    | 0.196  |
| Japanese | 0.236  | 0.252   | 0.248  | 0.272    | 0.249  | 0.225   | 0.000    | 0.108  |
| Korean   | 0.202  | 0.209   | 0.225  | 0.229    | 0.210  | 0.196   | 0.108    | 0.000  |

In previous research, we quantified a novel linguistic similarity metric from the WALS database based on the features all of the proposed languages shared. One of the problems of the metric was that as the amount of languages increased, the amount of features shared with them decreased due to missing values in the database. This time, we improved the calculation process and increased the robustness of the metric. The improved version attempts to counter the issue caused by the diminishing feature count. This was done by selecting all of the features that would have a defined value for both languages in all possible language pairs instead of having to be shared between all of the languages. The language pairs were formed from our proposed languages (English, German, Danish, Polish, Russian, Croatian, Japanese and Korean). Otherwise, the process remained the same. In short, we converted the possible feature values into numeric and calculated an average euclidean distance between all language pairs. This resulted in a symmetric distance metric. The goal was to create a multidomain similarity metric that would also be coherent and try to preserve the ordinality of the feature values. The finished distance matrix is shown in Table 4.

Lastly, our plan was to take a look at the STL similarity measure [22], which is based on multiple linguistic features. The measure puts together three different aspects of language by using Semantic, Terminological (lexical) and Linguistic (syntactic) similarity to form a single metric. According to the authors, the STL metric outperformed many previous measures that were relying only on one of the previously mentioned feature types [31, 32]. However, in order to be able to use the metric, the vocabulary dataset must be structured in the form of an ontology, which restricts the metric's use. Due to this fact and because of the lack of available languages for the used dataset, it was not possible for us to utilize the metric in this research.

## 5. Experiments

### 5.1. Setup

We fine-tuned both of the models (mBERT, XLM-R) with all of the proposed languages (English, German, Danish, Polish, Russian, Croatian, Japanese and Korean) for all of the tasks. Fine-tuning refers to the training of the parameters of a pre-trained language model (like BERT) with task-specific labeled data. This produced 16 fine-tuned models for each task, which sums up to a total of 48 fine-tuned models (two transformer models, eight languages, three tasks). After fine-tuning, we evaluated the models with test datasets from all of the previously mentioned languages to compute the cross-lingual zero-shot transfer scores. We did not use a train-dev-test, but only train-test scenario for evaluation, because the test dataset has nothing to do with the training dataset in a zero-shot task. We also do not aim at optimizing for each dataset, or creating a product, but rather study general properties. We evaluated the models with a macro F1-score for sentiment analysis and NER, and Label Attachment Score (LAS) for the dependency parsing task.

After finishing the evaluations for all of the fine-tuned models, we took a look at the correlation between the zero-shot cross-lingual transfer scores and linguistic similarity. This was done by using the four previously introduced linguistic similarity metrics (eLinguistics, EzGlot, WALS and lang2vec). We computed Pearson's and Spearman's correlations between the models' cross-lingual zero-shot transfer scores and the language similarity measures. The tasks, models and linguistic similarity metrics used in the experiments are listed in Table 5. The models were fine-tuned by using PyTorch and the Huggingface Transformers library [87]. The hardware used was an Nvidia GTX 1080Ti GPU.

**Table 5**
Tasks, models and linguistic similarity metrics used in the experiments

| Tasks | Models | Linguistic Similarity Metrics |
|---|---|---|
| Sentiment Analysis | mBERT | EzGlot |
| Named Entity Recognition | XLM-RoBERTa | eLinguistics |
| Dependency Parsing | | Averaged lang2vec |
| | | Quantified WALS |

**Table 6**
Sentiment analysis: F1-scores for mBERT

| | | **TARGET** | | | | | | | |
|---|---|---|---|---|---|---|---|---|---|
| | | Danish | English | German | Croatian | Polish | Russian | Japanese | Korean |
| | **Danish** | 0.976 | 0.875 | 0.951 | 0.941 | 0.934 | 0.876 | 0.800 | 0.881 |
| | **English** | 0.942 | 0.935 | 0.935 | 0.921 | 0.921 | 0.849 | 0.645 | 0.838 |
| | **German** | 0.901 | 0.816 | 0.971 | 0.908 | 0.889 | 0.828 | 0.711 | 0.741 |
| **SOURCE** | **Croatian** | 0.952 | 0.883 | 0.948 | 0.973 | 0.940 | 0.863 | 0.802 | 0.881 |
| | **Polish** | 0.952 | 0.876 | 0.948 | 0.948 | 0.967 | 0.861 | 0.771 | 0.878 |
| | **Russian** | 0.949 | 0.862 | 0.939 | 0.938 | 0.933 | 0.957 | 0.774 | 0.867 |
| | **Japanese** | 0.908 | 0.799 | 0.903 | 0.894 | 0.870 | 0.807 | 0.914 | 0.869 |
| | **Korean** | 0.940 | 0.848 | 0.935 | 0.930 | 0.909 | 0.850 | 0.815 | 0.957 |

**Table 7**
Sentiment analysis: F1-scores for XLM-R

| | | **TARGET** | | | | | | | |
|---|---|---|---|---|---|---|---|---|---|
| | | Danish | English | German | Croatian | Polish | Russian | Japanese | Korean |
| | **Danish** | 0.972 | 0.857 | 0.932 | 0.934 | 0.926 | 0.869 | 0.749 | 0.846 |
| | **English** | 0.939 | 0.925 | 0.920 | 0.922 | 0.916 | 0.844 | 0.705 | 0.816 |
| | **German** | 0.882 | 0.791 | 0.966 | 0.890 | 0.871 | 0.816 | 0.671 | 0.711 |
| **SOURCE** | **Croatian** | 0.945 | 0.859 | 0.925 | 0.969 | 0.929 | 0.876 | 0.763 | 0.835 |
| | **Polish** | 0.946 | 0.853 | 0.929 | 0.939 | 0.960 | 0.865 | 0.632 | 0.834 |
| | **Russian** | 0.918 | 0.792 | 0.895 | 0.913 | 0.901 | 0.953 | 0.726 | 0.832 |
| | **Japanese** | 0.888 | 0.793 | 0.880 | 0.871 | 0.851 | 0.773 | 0.905 | 0.832 |
| | **Korean** | 0.921 | 0.804 | 0.903 | 0.911 | 0.880 | 0.824 | 0.690 | 0.953 |

## 5.2. Results

Both of the multilingual transformer models (mBERT, XLM-R) were fine-tuned with all of the proposed languages for each task (sentiment analysis, NER, DEP) we introduced earlier. The models were fine-tuned using only the training dataset from a single language before the evaluation step. The model evaluation scores are presented in Tables 6 and 7 for sentiment analysis, Tables 8 and 9 for NER and Tables 10 and 11 for DEP.

Looking at the results, we can clearly say that XLM-R outperformed mBERT in all of the tasks. The only exception to this was the sentiment analysis task, where mBERT slightly outperformed XLM-R. It can be noted from the results that the highest transfer scores usually belong to the languages in the same language family as the source language (English, German, Danish - Germanic; Croatian, Polish, Russian - Slavic; Japanese, Korean - Koreano-Japonic). Also, most of the time there is a clear difference in the scores when evaluating with the same language as the source compared to zero-shot cross-lingual transfer. The exceptions to this are the sentiment analysis task for both models and the NER task for XLM-R.

**Table 8**
NER: F1-scores for mBERT

| | | | | | TARGET | | | | |
|---|---|---|---|---|---|---|---|---|---|
| | | **Danish** | **English** | **German** | **Croatian** | **Polish** | **Russian** | **Japanese** | **Korean** |
| | **Danish** | 0.957 | 0.813 | 0.801 | 0.480 | 0.763 | 0.791 | 0.675 | 0.640 |
| | **English** | 0.778 | 0.930 | 0.827 | 0.744 | 0.852 | 0.729 | 0.773 | 0.652 |
| | **German** | 0.770 | 0.866 | 0.936 | 0.751 | 0.879 | 0.805 | 0.766 | 0.648 |
| **SOURCE** | **Croatian** | 0.667 | 0.710 | 0.748 | 0.876 | 0.727 | 0.770 | 0.707 | 0.622 |
| | **Polish** | 0.691 | 0.676 | 0.702 | 0.695 | 0.956 | 0.648 | 0.754 | 0.625 |
| | **Russian** | 0.759 | 0.825 | 0.764 | 0.761 | 0.867 | 0.946 | 0.759 | 0.564 |
| | **Japanese** | 0.754 | 0.827 | 0.761 | 0.659 | 0.693 | 0.743 | 0.926 | 0.673 |
| | **Korean** | 0.602 | 0.694 | 0.668 | 0.670 | 0.675 | 0.705 | 0.700 | 0.867 |

**Table 9**
NER: F1-scores for XLM-R

| | | | | | TARGET | | | | |
|---|---|---|---|---|---|---|---|---|---|
| | | **Danish** | **English** | **German** | **Croatian** | **Polish** | **Russian** | **Japanese** | **Korean** |
| | **Danish** | 0.975 | 0.868 | 0.876 | 0.723 | 0.958 | 0.910 | 0.873 | 0.755 |
| | **English** | 0.955 | 0.941 | 0.941 | 0.820 | 0.961 | 0.934 | 0.920 | 0.781 |
| | **German** | 0.960 | 0.926 | 0.948 | 0.846 | 0.975 | 0.930 | 0.918 | 0.765 |
| **SOURCE** | **Croatian** | 0.920 | 0.842 | 0.872 | 0.911 | 0.919 | 0.889 | 0.834 | 0.723 |
| | **Polish** | 0.949 | 0.885 | 0.897 | 0.877 | 0.981 | 0.897 | 0.891 | 0.740 |
| | **Russian** | 0.923 | 0.908 | 0.915 | 0.611 | 0.951 | 0.951 | 0.880 | 0.737 |
| | **Japanese** | 0.955 | 0.909 | 0.920 | 0.847 | 0.961 | 0.926 | 0.936 | 0.789 |
| | **Korean** | 0.827 | 0.752 | 0.799 | 0.491 | 0.751 | 0.820 | 0.840 | 0.900 |

**Table 10**
DEP: LAS-scores for mBERT

| | | | | | TARGET | | | | |
|---|---|---|---|---|---|---|---|---|---|
| | | **Danish** | **English** | **German** | **Croatian** | **Polish** | **Russian** | **Japanese** | **Korean** |
| | **Danish** | 0.860 | 0.545 | 0.631 | 0.619 | 0.556 | 0.647 | 0.092 | 0.026 |
| | **English** | 0.652 | 0.891 | 0.670 | 0.624 | 0.570 | 0.653 | 0.165 | 0.021 |
| | **German** | 0.635 | 0.603 | 0.842 | 0.672 | 0.613 | 0.733 | 0.130 | 0.062 |
| **SOURCE** | **Croatian** | 0.581 | 0.607 | 0.633 | 0.893 | 0.645 | 0.778 | 0.124 | 0.030 |
| | **Polish** | 0.520 | 0.518 | 0.577 | 0.676 | 0.924 | 0.760 | 0.112 | 0.023 |
| | **Russian** | 0.594 | 0.604 | 0.643 | 0.730 | 0.666 | 0.878 | 0.131 | 0.020 |
| | **Japanese** | 0.132 | 0.148 | 0.163 | 0.114 | 0.117 | 0.126 | 0.926 | 0.033 |
| | **Korean** | 0.058 | 0.065 | 0.060 | 0.035 | 0.045 | 0.054 | 0.035 | 0.293 |

In dependency parsing, XLM-R slightly outperformed mBERT as expected. However, in the sentiment analysis task mBERT scored slightly higher than XLM-R overall, with both models scoring high across all language pairs. Some language pairs even achieving zero-shot cross-lingual transfer F-score of over 0.95. In this task, there seems not to be a clear pattern what kind of language pairs tend to yield higher performance. For example, Slavic languages seem to work better as sources for Danish compared to German languages in the case of both models. The scores are also similarly high across the board for the NER task with XLM-R, with the model being able to achieve very high scores with zero-shot transfer. The performance difference between mBERT and XLM-R is also more noticeable in the case of NER.

As can be seen from Table 12, both Japanese and Korean worked decently well as cross-lingual transfer sources for both sentiment analysis and NER tasks, even though being very different from the other languages used in the experiments as they are the only non Indo-European languages. However, in the case of DEP their performance is extremely low. Except for this case with DEP, all of the proposed languages seem to be quite equal as cross-lingual transfer sources in general. Interestingly, German, Croatian and Russian seem to perform slightly better overall compared to the other languages, especially with mBERT. A similar phenomenon was also experienced by Turc et al. [40] and in our previous research [26].

**Table 11**
DEP: LAS-scores for XLM-R

| | | | | | TARGET | | | | |
|---|---|---|---|---|---|---|---|---|---|
| | | Danish | English | German | Croatian | Polish | Russian | Japanese | Korean |
| | Danish | 0.888 | 0.679 | 0.725 | 0.706 | 0.672 | 0.715 | 0.095 | 0.366 |
| | English | 0.733 | 0.911 | 0.728 | 0.720 | 0.700 | 0.716 | 0.112 | 0.364 |
| | German | 0.712 | 0.681 | 0.854 | 0.751 | 0.732 | 0.784 | 0.066 | 0.405 |
| SOURCE | Croatian | 0.639 | 0.668 | 0.702 | 0.910 | 0.798 | 0.818 | 0.069 | 0.375 |
| | Polish | 0.614 | 0.603 | 0.676 | 0.780 | 0.945 | 0.804 | 0.049 | 0.384 |
| | Russian | 0.642 | 0.645 | 0.722 | 0.801 | 0.796 | 0.890 | 0.101 | 0.378 |
| | Japanese | 0.118 | 0.132 | 0.172 | 0.098 | 0.122 | 0.106 | 0.937 | 0.317 |
| | Korean | 0.326 | 0.292 | 0.381 | 0.298 | 0.325 | 0.304 | 0.183 | 0.877 |

**Table 12**
Average scores for each source language on each task

| | mBERT | | | XLM-R | | |
|---|---|---|---|---|---|---|
| | Sentiment | NER | DEP | Sentiment | NER | DEP |
| Danish | 0.904 | 0.740 | 0.497 | 0.886 | 0.867 | 0.606 |
| English | 0.873 | 0.786 | 0.531 | 0.874 | 0.907 | 0.623 |
| German | 0.845 | 0.803 | 0.536 | 0.825 | 0.909 | 0.623 |
| Croatian | 0.905 | 0.728 | 0.536 | 0.888 | 0.864 | 0.622 |
| Polish | 0.900 | 0.719 | 0.514 | 0.870 | 0.890 | 0.607 |
| Russian | 0.902 | 0.781 | 0.533 | 0.866 | 0.860 | 0.622 |
| Japanese | 0.871 | 0.754 | 0.220 | 0.849 | 0.905 | 0.250 |
| Korean | 0.898 | 0.698 | 0.081 | 0.861 | 0.773 | 0.373 |

## 5.3. Effect of Linguistic Similarity

We calculated the correlation between the zero-shot cross-lingual transfer results of the two models and each of the four proposed linguistic similarity metrics (EzGlot, eLinguistics, WALS and lang2vec) in all proposed NLP tasks using both Pearson's and Spearman's correlation coefficients ($p$-value). We were forced to ignore some of the language pairs when calculating the correlations with the EzGlot metric as some of the similarity values were missing. The correlation analysis results are shown in Table 13 for sentiment analysis, Table 14 for NER, Table 15 for DEP.

Looking at the results, one can say that there is mostly a strong correlation between lang2vec, WALS and eLinguistics metrics and the cross-lingual zero-shot transfer score, and a strong-moderate correlation between the EzGlot metric and the transfer scores for NER and DEP. In the case of sentiment analysis, the correlation is noticeably lower with XLM-R, staying at a moderate level with all of the linguistic similarity metrics. The correlation is strongest with the dependency parsing task with XLM-R, with the highest absolute Spearman's correlation being 0.897 with eLinguistics metric. Also, the results show p-value < 0.05 for all of the tasks, models and metrics, indicating statistical significance.

For both of the models, the correlation for lang2vec, WALS and eLinguistics metrics are generally higher than EzGlot, except in the case of sentiment analysis, where EzGlot's correlation is slightly higher than WALS and eLinguistics using both Pearson's and Spearman's correlation coefficients. Also, the correlations were generally slightly stronger with mBERT in sentiment analysis, while XLM-R had higher correlations in both NER and DEP tasks.

However, the results changed drastically for all tasks except dependency parsing, when we removed the anchor points of same source-target language pairs (monolingual scenarios), leaving only the zero-shot transfer results. This was necessary to do in order remove the bias brought by the monolingual data points, as the scores are higher and the languages would also be the most similar (same). The results after removing the same source-target language pairs are shown in Table 16 for sentiment analysis, Table 17 for NER, Table 18 for DEP.

**Table 13**
Sentiment analysis: Pearson's and Spearman's correlation coefficients for model F1 scores and linguistic similarity metrics

| | Pearson | | | | Spearman | | | |
|---|---|---|---|---|---|---|---|---|
| | XLM-R | | mBERT | | XLM-R | | mBERT | |
| | $\rho$ | p-value | $\rho$ | p-value | $\rho$ | p-value | $\rho$ | p-value |
| WALS | -0.297 | 0.017 | -0.645 | 0.000 | -0.331 | 0.008 | -0.537 | 0.000 |
| EzGlot | 0.389 | 0.005 | 0.729 | 0.000 | 0.533 | 0.000 | 0.586 | 0.000 |
| eLinguistics | -0.355 | 0.004 | -0.648 | 0.000 | -0.413 | 0.001 | -0.652 | 0.000 |
| lang2vec | -0.418 | 0.001 | -0.746 | 0.000 | -0.482 | 0.000 | -0.623 | 0.000 |

**Table 14**
NER: Pearson's and Spearman's correlation coefficients for model F1 scores and linguistic similarity metrics

| | Pearson | | | | Spearman | | | |
|---|---|---|---|---|---|---|---|---|
| | XLM-R | | mBERT | | XLM-R | | mBERT | |
| | $\rho$ | p-value | $\rho$ | p-value | $\rho$ | p-value | $\rho$ | p-value |
| WALS | -0.514 | 0.000 | -0.500 | 0.000 | -0.510 | 0.000 | -0.486 | 0.000 |
| EzGlot | 0.494 | 0.000 | 0.427 | 0.002 | 0.464 | 0.001 | 0.401 | 0.004 |
| eLinguistics | -0.580 | 0.000 | -0.517 | 0.000 | -0.608 | 0.000 | -0.553 | 0.000 |
| lang2vec | -0.465 | 0.000 | -0.432 | 0.000 | -0.504 | 0.000 | -0.461 | 0.000 |

**Table 15**
DEP: Pearson's and Spearman's correlation coefficients for model LAS scores and linguistic similarity metrics

| | Pearson | | | | Spearman | | | |
|---|---|---|---|---|---|---|---|---|
| | XLM-R | | mBERT | | XLM-R | | mBERT | |
| | $\rho$ | p-value | $\rho$ | p-value | $\rho$ | p-value | $\rho$ | p-value |
| WALS | -0.781 | 0.000 | -0.718 | 0.000 | -0.844 | 0.000 | -0.693 | 0.000 |
| EzGlot | 0.588 | 0.000 | 0.516 | 0.000 | 0.694 | 0.000 | 0.561 | 0.000 |
| eLinguistics | -0.845 | 0.000 | -0.840 | 0.000 | -0.897 | 0.000 | -0.867 | 0.000 |
| lang2vec | -0.702 | 0.000 | -0.679 | 0.000 | -0.848 | 0.000 | -0.754 | 0.000 |

First of all, for eLinguistics and WALS similarity metrics, the correlations generally dropped from strong to moderate for the NER task while EzGlot's correlation fell close to zero and also lost all statistical significance. The correlation of lang2vec also plummeted and mostly lost statistical significance, but remained higher than EzGlot's. Interestingly, for sentiment analysis, the result looks completely opposite. The correlations of EzGlot and lang2vec only fell slightly while WALS' and eLinguistics' correlation plummeted down and lost statistical significance for XLM-R. The correlations in the dependency parsing task only dropped slightly for all of the linguistic similarity metrics. Also, after removing the same source-target language pairs, the strongest correlations are still found in DEP, followed by NER, while sentiment analysis still has the weakest correlations overall. However, as linguistic similarity still correlates with cross-lingual transfer performance, we can get an improved model performance by using linguistic similarity for transfer language selection.

**Table 16**
Sentiment analysis: Pearson's and Spearman's correlation coefficients for model F1 scores and linguistic similarity metrics for zero-shot only

| | Pearson | | | | Spearman | | | |
| | XLM-R | | mBERT | | XLM-R | | mBERT | |
| | $\rho$ | p-value | $\rho$ | p-value | $\rho$ | p-value | $\rho$ | p-value |
|---|---|---|---|---|---|---|---|---|
| **WALS** | -0.111 | 0.415 | -0.262 | 0.051 | -0.168 | 0.216 | -0.315 | 0.018 |
| **EzGlot** | 0.327 | 0.035 | 0.303 | 0.051 | 0.403 | 0.008 | 0.313 | 0.044 |
| **eLinguistics** | -0.229 | 0.090 | -0.392 | 0.003 | -0.284 | 0.034 | -0.487 | 0.000 |
| **lang2vec** | -0.380 | 0.004 | -0.454 | 0.000 | -0.374 | 0.005 | -0.440 | 0.001 |

**Table 17**
NER: Pearson's and Spearman's correlation coefficients for model F1 scores and linguistic similarity metrics for zero-shot only

| | Pearson | | | | Spearman | | | |
| | XLM-R | | mBERT | | XLM-R | | mBERT | |
| | $\rho$ | p-value | $\rho$ | p-value | $\rho$ | p-value | $\rho$ | p-value |
|---|---|---|---|---|---|---|---|---|
| **WALS** | -0.336 | 0.011 | -0.347 | 0.009 | -0.316 | 0.017 | -0.293 | 0.028 |
| **EzGlot** | 0.173 | 0.274 | 0.120 | 0.448 | 0.170 | 0.282 | 0.095 | 0.549 |
| **eLinguistics** | -0.453 | 0.000 | -0.384 | 0.003 | -0.458 | 0.000 | -0.389 | 0.003 |
| **lang2vec** | -0.234 | 0.082 | -0.214 | 0.113 | -0.307 | 0.021 | -0.256 | 0.057 |

**Table 18**
DEP: Pearson's and Spearman's correlation coefficients for model LAS scores and linguistic similarity metrics for zero-shot only

| | Pearson | | | | Spearman | | | |
| | XLM-R | | mBERT | | XLM-R | | mBERT | |
| | $\rho$ | p-value | $\rho$ | p-value | $\rho$ | p-value | $\rho$ | p-value |
|---|---|---|---|---|---|---|---|---|
| **WALS** | -0.738 | 0.000 | -0.661 | 0.000 | -0.769 | 0.000 | -0.588 | 0.000 |
| **EzGlot** | 0.421 | 0.005 | 0.373 | 0.015 | 0.488 | 0.001 | 0.349 | 0.024 |
| **eLinguistics** | -0.795 | 0.000 | -0.822 | 0.000 | -0.848 | 0.000 | -0.842 | 0.000 |
| **lang2vec** | -0.714 | 0.000 | -0.709 | 0.000 | -0.775 | 0.000 | -0.676 | 0.000 |

## 6. Discussion

### 6.1. Transfer Language Performance

XLM-R outperforming mBERT generally matches our expectations, as it also did so on a variety of benchmark tasks [12, 17]. The reason behind this most likely is the fact that XLM-R uses a vastly larger amount of data for pretraining compared to mBERT. The performance difference between the two models is the most clear in the NER task.

According to the results, simply choosing English as the transfer source did not yield top results most of the time, sometimes even as a source language to other languages in the Germanic language group. For example, it had a lower than average performance in sentiment analysis and an average performance in the other two tasks probably due to its simplicity when compared to both Danish and German. It was also slightly outperformed by Slavic languages in some cases when used as a source for other Germanic languages. Another reason could be the influence of French [88, 89, 90], which might further distance it from the other Germanic languages. Also, the differences in morphology could be a factor here. Danish and German probably work better with each other due to a great amount of historic mutual influence.

In sentiment analysis, the models achieved slightly better scores with English and it was on-par with other Germanic languages. However, all of the Slavic languages still tended to work slightly better as transfer sources. These results show that other languages should also be considered over English as the cross-lingual transfer source if available. For the other two tasks, NER and DEP, English performed well and showed to be a good transfer source for the other two Germanic languages.

In most cases, using languages from the same language family as the source language yielded the highest cross-lingual transfer scores. This matches with the typical intuition-based selection process used to select source language for cross-lingual transfer. However, relying only on intuition and looking purely at language families when when selecting the transfer language will lead to diminished results in some cases.

One example would be taking Polish as the target language for NER task. One could expect that in this case, the best transfer languages would be Croatian and Russian, but looking at the results (Tables 8 and 9) German had a higher cross-lingual transfer score even though it is from the Germanic language family, not Slavic. This could be, for example, due to mutual influence of these two languages. The grammar of both Danish and English is relatively simple compared to German, which could aid them in generalizing better with one another. Looking at the scores, it can be noted that German is a good source for both Germanic and Slavic languages, which could mean that the historical mutual influence between the Germans and Slavs could be a factor here. Furthermore, German, in addition to having a higher average performance on most tasks, tended to also work exceptionally well as a source language for other Slavic languages, most likely because of the reasons discussed above.

In addition, Japanese and Korean did not achieve comparably better scores with one another, contrary to our expectations, and were even slightly outperformed by the other languages approximately half of the time, even though being more similar with each other compared to any of the other proposed languages. Here the reason could be, for example, the differences in the writing systems, as neither of these two languages use alphabets and their systems also greatly differ from each other.

Also, both Russian and Croatian had a higher than average performance on most of the tasks. This was similar to our previous research [26] where Russian performed exceptionally well as a transfer source for offensive language identification. However, unlike in our previous research, Russian did not perform noticeably well as the transfer language source for Korean and Japanese. Thus the phenomenon experienced previously is most likely related to the topic of offensive language identification itself or to the properties of these specific datasets. We will investigate this in later research. Also, Japanese and Korean had a satisfying performance as source languages for the Germanic and Slavic languages in both sentiment analysis and NER tasks, even though Japanese and Korean are fundamentally different from the languages of these two families, as they are the only non Indo-European languages in the proposed set. This demonstrates that multilingual transformer models are also able to leverage knowledge even from very distant languages.

## 6.2. Analysis of Linguistic Similarity Metrics

The correlation between cross-lingual transfer performance and the similarity metrics were strong or moderate with all of the proposed metrics, which would suggest that using even a single feature such as lexical information or by comparing phonetic consonants is still effective to some extent.

### 6.2.1. EzGlot

However, when considering only the zero-shot transfer results, EzGlot's similarity metric's correlation dropped drastically and out of statistical significance in the NER task. This shows that it does not necessarily rely on lexical features and that other linguistic features need to be considered when choosing the source language for NER. On the other hand, the same happened with the lang2vec metric despite it being created using different features from multiple domains. Also, the opposite happened in the sentiment analysis task, as both WALS and eLinguistics metrics' correlation dropped drastically and out of statistical significance. This hints the importance of lexical similarity when choosing the source language for sentiment analysis tasks.

### 6.2.2. eLinguistics

Surprisingly, even though using only a predefined set of phonetic consonants for its calculation, the correlation of eLinguistics' similarity metric was stronger in all tasks compared to the the correlation of the WALS metric, which we quantified from the WALS database using linguistic features from multiple domains. The correlation of eLinguistics was also higher than the averaged lang2vec metric in both NER and DEP. The reason behind this could be that including all of possible features between each language pair could have caused too many irrelevant features to be included. This can cause a possible bias the metric calculation.

However, the eLinguistics metric also has its weak points as it is based on only a single aspect of language, even though its correlation being the strongest. One can see from Table 1 that eLinguistics shows Japanese being very distant from Korean, being at the same level as Polish and Russian, with Croatian being seemingly closer to Korean than Japanese, which is not true due to the similarities in the vocabulary and grammar of Japanese and Korean. Taking a look at Tables 4 and 3, it is clear that the WALS and lang2vec metrics are a lot more robust to this kind of errors. The reason most likely is that instead of using only a single linguistic feature like the eLinguistics metric, the WALS and lang2vec metrics are based on a large amount of features spanning over multiple domains.

### 6.2.3. Averaged lang2vec and Quantified WALS

Both lang2vec and WALS had a strong correlation with the DEP task. Although both metrics are based on a large number of linguistic features spanning over multiple domains, their correlations varied greatly with the other two tasks. Specifically, in NER, the correlation of lang2vec was noticeably lower. In sentiment analysis, the correlation of WALS plummeted while lang2vec stayed at a moderate level. The reason is probably in the calculation of the metrics. In the quantified WALS metric, features are treated as continuous whereas lang2vec uses one-hot encoding. Also, in quantified WALS, every single feature has the same weight whereas in averaged lang2vec every category of features has the same weight but might contain a different amount of features. Additionally, the features in lang2vec are collected from multiple sources, which increases the amount of data while possibly introducing incoherence. Lastly, the number of features used in the similarity calculations with the quantified WALS metric varies slightly between language pairs due to missing values in the database. Lang2vec tries to counter this by using a model to predict the missing values, although this might introduce more errors in some cases.

In the future, we will take another glance at the WALS database, aiming for a better quantification by looking at the importance of each feature group (syntactic, lexical, phonetic, etc.) and weighing accordingly while filtering out redundant features in order to develop an even more effective and comprehensive similarity metric. We will also take a look at lang2vec, aiming to filter out redundant features and weigh the categories accordingly instead of simply taking an average in order to make it better suited for transfer language selection. After all, both WALS and lang2vec metrics are more robust thanks to being based on a large amount of features spanning over multiple domains instead of using only a single linguistic feature like eLinguistics or EzGlot.

## 6.3. Task-Specific Analysis
### 6.3.1. Sentiment Analysis

Looking at the f-scores of the sentiment analysis task, it is clear that the results are very high across the board and the score differences between language groups are also very small, with sometimes languages from other language groups than the target emerging as the best performers. This is the case for example with Danish, as Croatian achieved the highest zero-shot transfer scores for both mBERT and XLM-R instead of another Germanic language.

A trait only observed in this task was that lang2vec and EzGlot were the only metrics keeping a moderate correlation in the zero-shot setting. The fact that Ezglot's correlation stayed moderate hints the importance of lexical features. One could argue that the reason behind the overall high scores might be due to the task being too easy, as it simply required the classification of the entries into positive and negative. This has also been shown in other research [91]. However, this also shows that it could be possible to achieve at least close to state-of-the-art results with multilingual transformer models in a zero-shot cross-lingual setting. This raises questions about how to improve the cross-lingual models to better utilize cross-lingual transfer. In the future, it would be useful to further investigate the models' behaviour in zero-shot setting. This could also be useful in the further development of measures to support low-resource languages.

### 6.3.2. Named Entity Recognition

For mBERT, the zero-shot results of the NER task look clearly lower than with same language pairs and quite even across all of the proposed languages and the languages belonging to the same group having generally a slightly higher score. However, the results of XLM-R closely resemble those of the sentiment analysis task as the results are considerably high across all language pairs. This further shows the potential these models have in relieving the issues with low-resource languages. Also, there is a moderate correlation between the zero-shot transfer performance and linguistic similarity for both WALS and eLinguistics metrics, and a weak/moderate correlation with lang2vec.

### 6.3.3. Dependency Parsing

The zero-shot results in DEP also seem clearly lower than with same language pairs and somewhat even across the board. The languages belonging to the same language family also generally have a slightly higher score. As the task requires the understanding of syntax and grammar and the scores are still reasonably high overall, the results could also support studies claiming that cross-lingual transformer models are able to learn grammar without explicit information [92].

On the other hand, Japanese and Korean had very poor performance as source languages while also being very difficult target languages in the DEP task, unlike in the sentiment analysis and NER tasks. This could mean that the model is unable to generalize to the syntax and grammar of Indo-European languages with Japonic-Koreanic languages and vice versa. The reason might be due to the differences in writing systems. In the DEP task, both XLM-R and mBERT keep a strong correlation between the zero-shot transfer performance and linguistic similarity with WALS, eLinguistics and lang2vec metrics and a moderate correlation with EzGlot in a zero-shot setting.

## 6.4. Impact

As there is a correlation with linguistic similarity and cross-lingual transfer performance for all of the tasks, including the abusive language identification task used in our previous research [26], it is possible to use linguistic similarity for transfer language selection, at least for these tasks. However, the correlation varied greatly from task to task, which means there is a lot of room for improvement in developing an optimal similarity metric for transfer language selection.

In order to confirm the efficacy of choosing another language over English as the cross-lingual transfer source, we performed a z-test between the results of using English as transfer source and using the language with the highest score. The test showed $Z = -3.18 < -1.96$ and $p = 0.001 < 0.05$ meaning that there is a significant difference between using English and the optimal language as the cross-lingual transfer source.

Based on these results, as there is a significant difference between using English and the optimal language as the cross-lingual transfer source, it is better to look for high-resource languages that have proper data available and are as close as possible to the target language based on a similarity metric instead of making a decision based on intuition or simply relying on English. This allows one to make a more informed and effective decision and makes model development more efficient.

## 6.5. Future Research

In the future, we are planning to analyze, what kind of linguistic features are the most important from the point of view of cross-lingual transfer. A solution could be grouping the features presented in WALS into syntactic, lexical, phonetic, etc., and calculating, which feature group has the strongest correlation with the cross-lingual transfer performance. We could then re-quantify the WALS database using this information in order to develop an even more effective and comprehensive similarity metric. It would also be beneficial if the WALS project received more attention and the feature matrix became more densely populated. Also, instead of taking an average of lang2vec's categories, they should be weighed by importance.

As shown by the DEP task, the models might be able to learn syntax and grammar without any explicit information. This could mean that adding explicit syntactic and grammatical information to the pre-training process of the models might also improve their performance. We will take a look at this in the future. Also, as the models achieved zero-shot transfer scores rivaling those of the monolingual settings, especially in sentiment analysis, it would be useful to perform an in-depth investigation about the models' behaviour in a zero-shot transfer learning setting to possibly find insights on how to improve their transfer learning capabilities.

## 7. Conclusions

In this research we studied cross-lingual transfer language selection for zero-shot learning using three different NLP tasks, namely, sentiment analysis, NER, and dependency parsing. We showed the effectiveness of cross-lingual zero-shot transfer learning with a total of eight languages from three language families. In this way, existing data from higher-resource languages may be used to improve the performance of languages that lack sufficient data.

We found a strong correlation between the similarity of the used languages and cross-lingual transfer performance. The transfer performance declines when the distance between the languages increases. This allows for the selection of a more suitable transfer language by assessing linguistic similarities rather than simply depending on intuition or defaulting to English. As our experiments have demonstrated, there is a significant difference in choosing the optimal transfer language over defaulting to English. As there is a correlation between linguistic similarity and transfer performance and a significant difference between using English and the optimal language as the cross-lingual transfer source, one should instead choose the source language based on a linguistic similarity measure. Our experiments also demonstrated that lexical information alone is insufficient to determine the optimal transfer languages at least for the tasks of NER and DEP.

it is better to look for high-resource languages that have proper data available and are as close as possible to the target language based on a similarity metric instead of making a decision based on intuition or simply relying on English. This allows one to make a more informed and effective decision and makes model development more efficient.

The results showed that the proposed method for cross-lingual transfer language selection could also be useful as a general method for other Natural Language Processing tasks, at least based on these tasks and our previous research. We also showed that it is possible to achieve good performance on the target language in a zero-shot cross-lingual transfer setting with multiple NLP tasks. This helps in developing better systems, especially when dealing with low-resource languages.

We improved a novel linguistic similarity metric consisting of various linguistic features by using the WALS database. Our proposed method did not show the strongest correlation with the transfer performance, but it still showed potential as a metric that could be useful for the selection process, especially if given a more refined or inclusive feature set. In the future, we will reassess the importances of the linguistic features used in the similarity metric calculation in order to have a more refined feature set, aiming to create an even more effective and comprehensive linguistic similarity metric.

Lastly, even though the overall high scores in the sentiment analysis task might be caused by the task being too easy, it also shows that it could be possible to achieve results close to those of a monolingual fine-tuning in a zero-shot cross-lingual transfer setting. This means it could be useful to thoroughly investigate the models' behaviour in zero-shot setting in order to find insights to improving their transfer capabilities. Also, as the DEP task demonstrated that the models might have a capability to understand grammar, adding explicit syntactic and grammatical information to the models' pre-training could also increase performance.

## CRediT authorship contribution statement

**Juuso Eronen:** Conceptualization, Methodology, Software, Writing - Original Draft, Writing - Review & Editing. **Michal Ptaszynski:** Conceptualization, Methodology, Supervision, Data Curation. **Fumito Masui:** Conceptualization, Methodology, Supervision, Data Curation.